\documentclass[10pt,twocolumn,letterpaper]{article}
\pdfoutput=1
\usepackage{cvpr}
\usepackage{times}
\usepackage{epsfig}
\usepackage{graphicx}
\usepackage{amsmath}
\usepackage{amssymb}
\graphicspath{ {./} }
\usepackage{authblk}
\usepackage[T1]{fontenc}
\usepackage[utf8]{inputenc}

\cvprfinalcopy 

\def\cvprPaperID{****} 

\ifcvprfinal\fi
\begin{document}

\title{The iMet Collection 2019 Challenge Dataset}


\author[1]{Chenyang Zhang}
\author[1]{Christine Kaeser-Chen}
\author[5]{Grace Vesom}
\author[3]{Jennie Choi}
\author[3]{Maria Kessler}
\author[1,3]{Serge Belongie}
\affil[1]{Google Research}
\affil[2]{The Metropolitan Museum of Art}
\affil[3]{Cornell University and Cornell Tech}
\affil[5]{Arraiy, Inc.}

\maketitle

\begin{abstract}
Existing computer vision technologies in artwork recognition focus mainly on instance retrieval or coarse-grained attribute classification. In this work, we present a novel dataset for fine-grained artwork attribute recognition. The images in the dataset are professional photographs of classic artworks from the Metropolitan Museum of Art, and annotations are curated and verified by world-class museum 
experts. In addition, we also present the iMet Collection 2019 Challenge as part of the FGVC6 workshop. Through the competition, we aim to spur the enthusiasm of the fine-grained visual recognition research community and advance the state-of-the-art in digital curation of museum collections.
\end{abstract}

\section{Introduction}
Fine grained visual categorization (FGVC) has emerged as a popular research area in computer vision in recent years. With the rapid progress of deep neural networks, computer vision algorithms can now capture powerful representations for complex semantics in domains such as fashion and biodiversity. Large-scale datasets such as COCO \cite{lin2014microsoft} and ImageNet  \cite{deng2009imagenet} have played important roles in helping advance state-of-the-art algorithms for coarse-grained category recognition, and the emergence of FGVC datasets provide a complementary role for a wide variety of subcategories.

\begin{figure}[t]
    \centering
    \includegraphics[width=1.0\columnwidth]{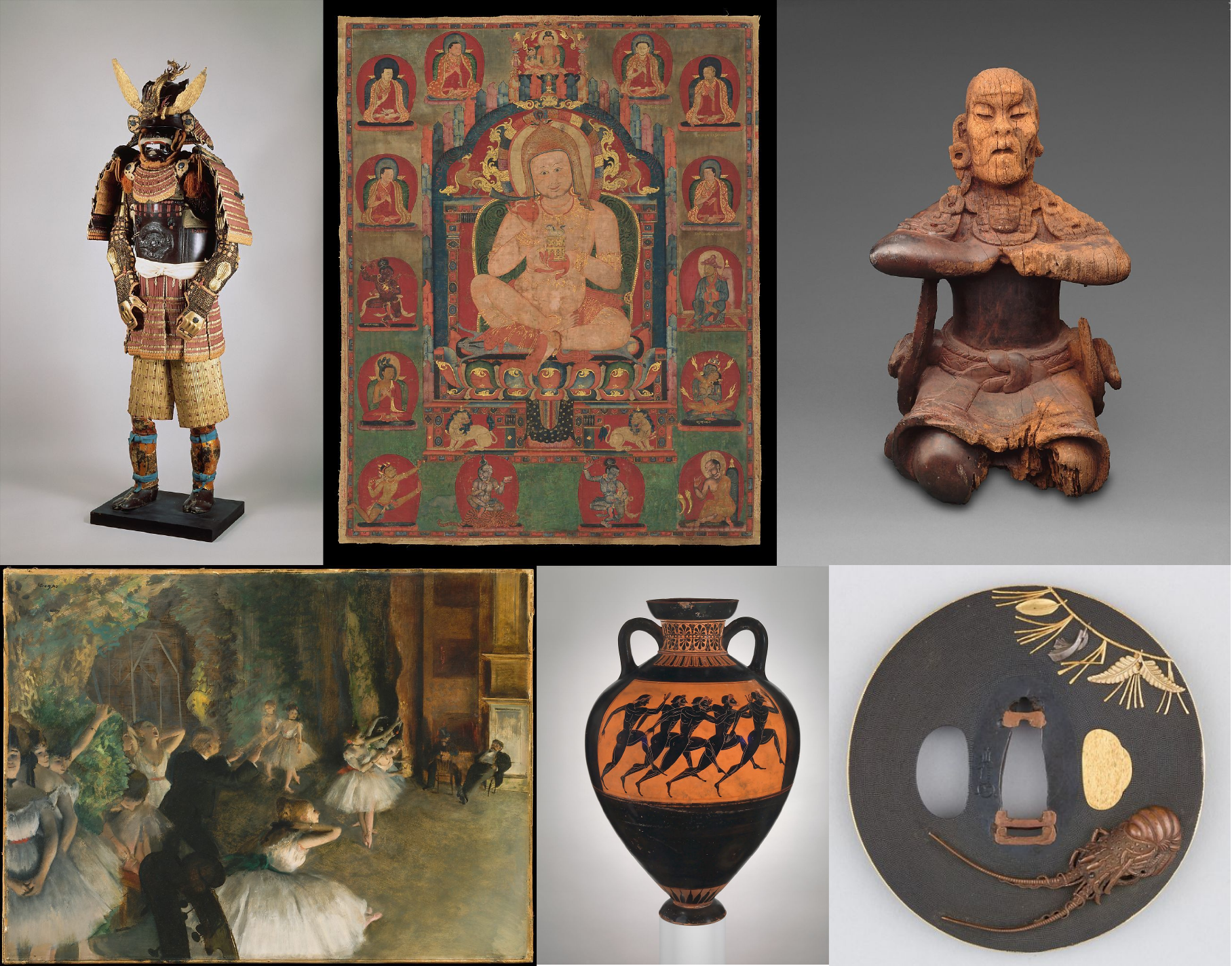}
    \caption{Sample images from the iMet Collection dataset.}
    \label{fig:showcase}
    \vspace{-5pt}
\end{figure}

Previous works on artwork recognition focus mainly on instance retrieval \cite{crowley2014search,gorisse20073d}, mid-level attributes such as color \cite{yelizaveta2005analysis}, and person detection \cite{ginosar2014detecting}. Other datasets feature descriptive attributes such as style, theme and mood \cite{krasin2017openimages,wilber2017bam}.

In this work, we present a novel dataset, \textit{iMet Collection 2019}, focusing on fine-grained artwork attribute recognition.
Compared to previous artwork datasets, the proposed dataset features the following two characteristics:
\vspace{-5pt}
\begin{itemize}
    \item \textbf{Fine-grained} The proposed dataset contains diverse attributes from a specialized domain. 
    \vspace{-2pt}
    \item \textbf{Research-grade} Museum experts curated and verified attribute labels to ensure high quality.
\end{itemize}

\begin{figure}[htb!]
    \centering
    \includegraphics[width=1.0\columnwidth]{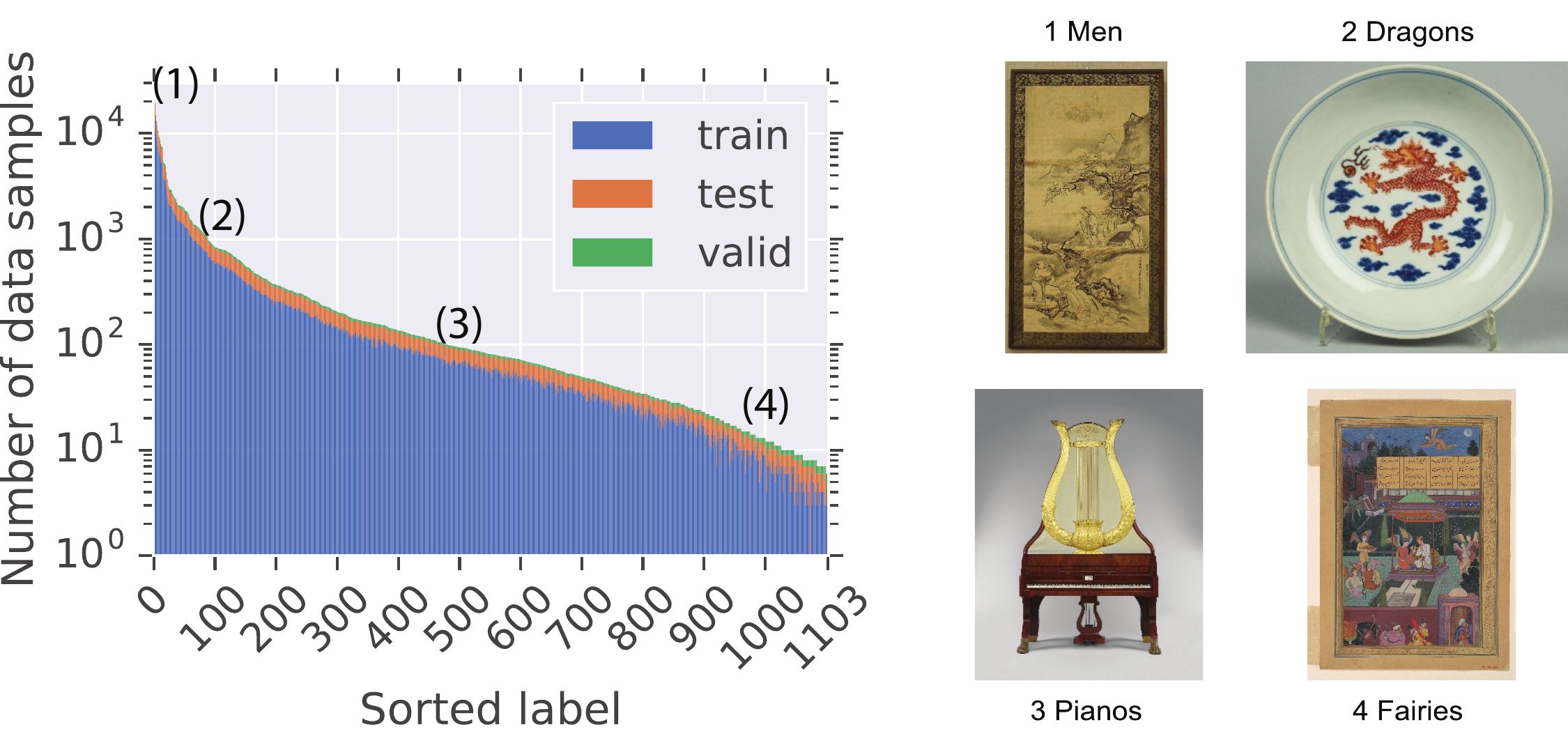}
    \caption{Left: distribution of images per attribute label. Right: representative images for 4 sampled attributes.}
    \label{fig:sample_label}
    \vspace{-10pt}
\end{figure}

\vspace{-10pt}
\section{The iMet Collection Dataset}
The Metropolitan Museum of Art in New York (The Met) has a diverse collection of artworks from across the globe spanning 5,000 years.  The artworks in The Met’s collection include paintings, musical instruments, prints, arms and armor, clothing, sculpture, furniture, metalwork, and more.

In 2017 The Met launched its Open Access Program making over $375,000$ artwork images and data in its collection available under Creative Commons CC0 license.  The Open Access data includes object type, artist, title, period, creation date, medium, culture, dimensions, and geographic information. The Met recently added subject keyword tags to their Open Access Program to improve discoverability and increase access to the collection. Together the data and the subject keyword tags can also be used for content suggestion and a better means to navigate the collection.

To scale the coverage of the subject keyword tags to all artwork images, we might need to leverage more automated annotation algorithms.  To that end, we release the iMet Collection dataset consisting of images with groundtruth keyword labels, which could facilitate the research in computer vision algorithms for keyword annotation as well as other applications.

\subsection{Images}
All images from our dataset are supplied by The Met and are all publicly available under CC0 license.\footnote{The Met Online Collection: https://www.metmuseum.org/art/collection} While there may be more than one image available for each \textit{object} (artwork instance), we only retain the ones for ``main-display.''\footnote{\textit{Main-display} refers to images capturing the most comprehensive view, often frontal, of the object.} Each image is resized such that the shorter dimension is 300 pixels and is encoded in PNG format. Sample images are shown in Figure \ref{fig:showcase}.

\subsection{Annotations}
The Met’s subject keyword tags cover religious subjects, historical events, plants and animals.  The tag set provides a new perspective into the collection, allowing users to search common topics across time.

The annotations of the dataset come from two sources: (1) Subject Matter Experts (SME) and (2) vendor annotators.

\textbf{SME derived attributes} are provided at curation time by the SME, and may contain single or multiple labels. These labels are drawn from several predefined domains such as \textit{Culture, Medium, Origin, etc}. For this competition, we only include the \textit{Culture} domain for simplicity.

\textbf{Vendor-labeled attributes}
were added by a professional outsourced organization from a taxonomy provided by The Met. The annotators were able to view the museum’s online collection pages and were advised to avoid annotating labels already present. Specifically, the vendor was advised to annotate labels which can be visually inferred (\textit{e.g.} ``Shellfish'' in Figure \ref{fig:showcase}, bottom-right sword guard). These annotations can be potentially more intuitive and visually plausible.

We have included $\mathbf{1103}$ different attributes from both culture-related attributes from SME derived attributes and vendor-sourced attributes in the final dataset.

\subsection{Data Splits and Sizes}
We divide the dataset into three splits: \textit{training}, \textit{validation}, and \textit{test} subsets. While the training and validation splits were released during the competition, we retained the test split for final ranking and will release it afterwards.

Since each data sample is annotated with 1 or more attributes, we first randomly pick a ``pivot'' attribute label for each sample. Then we split the dataset using the ``pivot'' attribute labels in the following manner:
\begin{enumerate}
    \item We discard samples whose pivot attributes have fewer than three samples.
    \vspace{-3pt}
    \item If an pivot attribute has fewer than 10 samples, we assign one to each of validation and test set, and the rest to training set.
    \vspace{-3pt}
    \item Otherwise, we assign $70\%$, $5\%$, and  $25\%$ of them to training, validation, and test, respectively.
    \vspace{-3pt}
\end{enumerate}
Finally, we reattach the non-pivot labels back to each assigned sample and discard labels which do not exist in all three splits.
In the end, our dataset contains $1103$ attribute labels from $155,531$ samples, among which we have $109,274$ samples for training, $7443$ for validation and $38,814$ for testing.

\subsection{Data Challenges}
\begin{figure}[htb!]
    \centering
    \includegraphics[width=0.8\columnwidth]{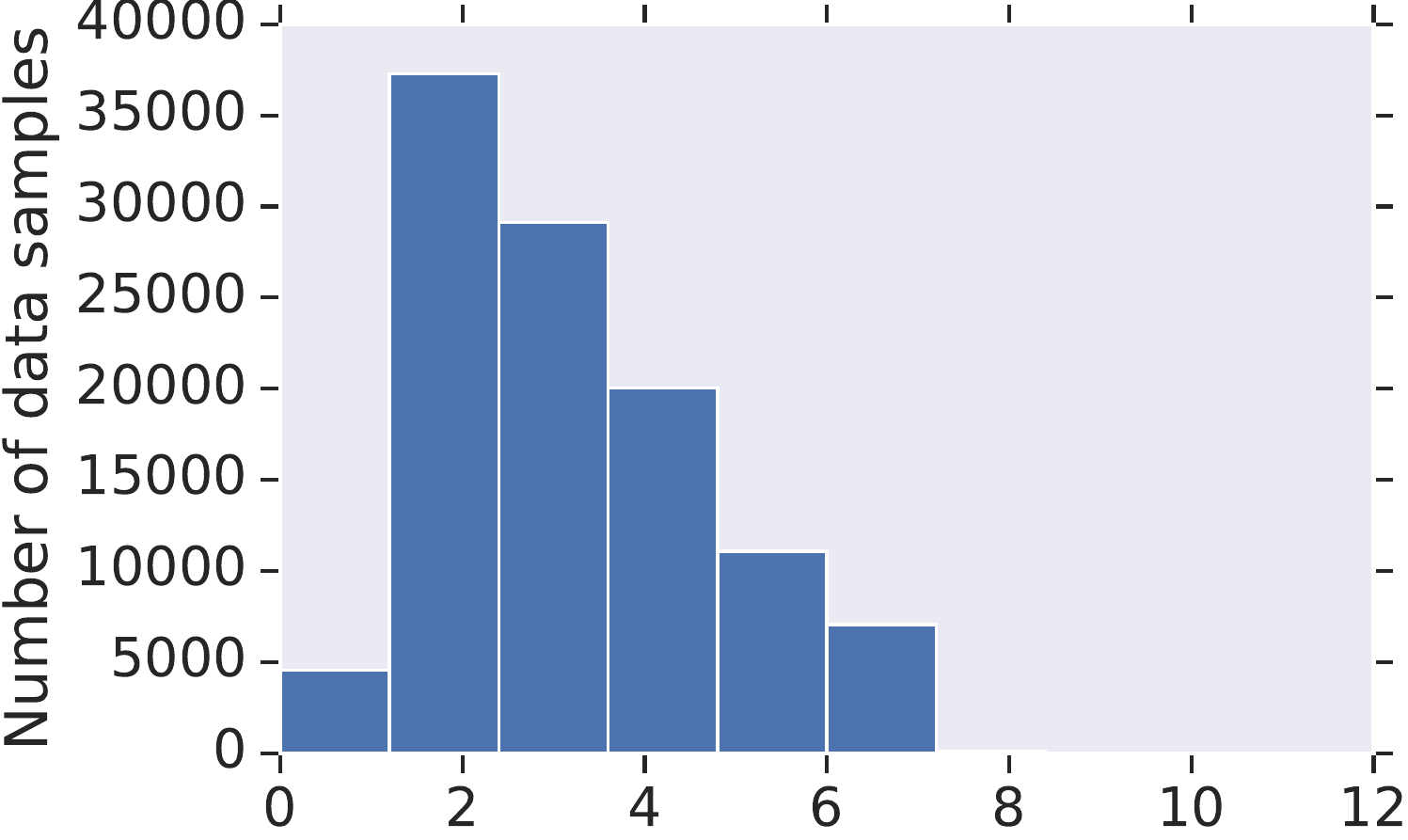}
    \caption{Histogram of per-sample attribute numbers over the training set. Most samples have 2 to 4 attributes while one of the training sample has 11 attributes.}
    \label{fig:attribute}
    \vspace{-5pt}
\end{figure}

The iMet Collection 2019 Dataset has a long-tailed distribution (Figure \ref{fig:sample_label}) over the attribute label set. The three most popular attributes are ``Men'', ``Women'', and ``French'' (which contain $19,974$, $14,283$, and $13,549$ training samples, respectively), and the least popular tag is ``Afgan'', having only one training sample. Another challenge is the unbalanced numbers of attributes per sample. As shown in Figure \ref{fig:attribute}, the number of attributes per sample ranges between $1$ and $11$, while most of the samples have $2$ to $4$ attributes.

\subsection{Metric}
We evaluate the performances of our dataset following the multi-label classification manner and employ the $F_\beta$ \cite{powers2011evaluation} score:
\vspace{-2pt}
\begin{equation}
\small
    F_{\beta} = \frac{(1+\beta^2)TP}{(1+\beta^2)TP + \beta^2 FN + FP}
\end{equation}
where TP, FN, and FP stand for numbers of \textit{true positives}, \textit{false negatives} and \textit{false positives}, respectively. In the challenge, we set $\beta=2$ to trade for more true positives with more tolerance to false positives.
\section{The iMet Collection Challenge 2019}
\begin{figure}[h]
    \centering
    \includegraphics[width=1.0\columnwidth]{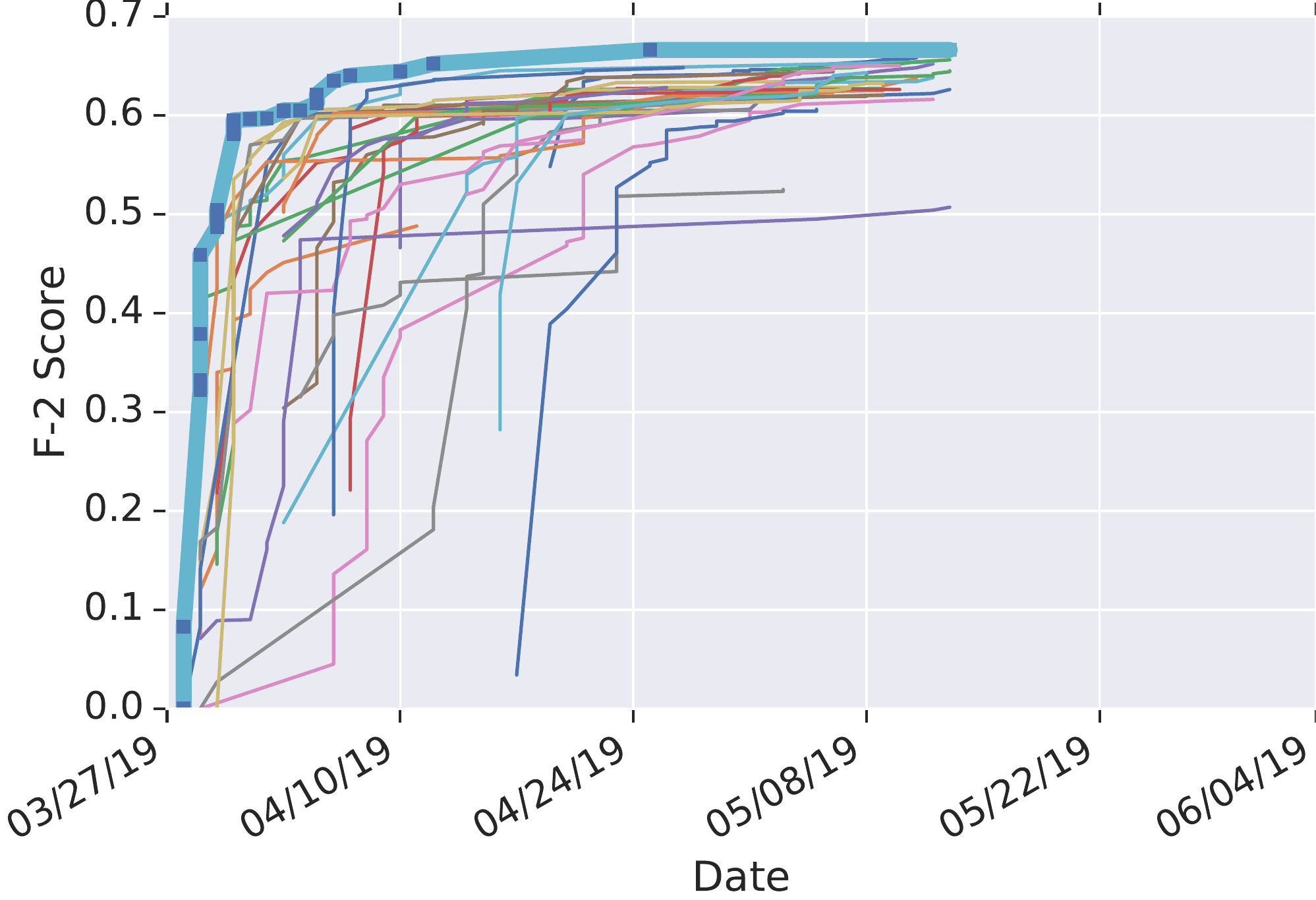}
    \caption{Challenge leader-board scores up to date. Each colored thin line represents a team with their submissions evaluated on validation set. The cyan thick curve indicates the leading score over time among all teams. We only plot 32 teams who have the longest date spans.}
    \label{fig:submission}
    \vspace{-5pt}
\end{figure}

The iMet Collection Challenge 2019 is conducted through Kaggle as part of FGVC6 workshop at CVPR19. At the time of paper submission, over \textbf{500} competitors from \textbf{435} teams have submitted their results. We plot the public leaderboard progress over time in Figure \ref{fig:submission}.
The competition is the first kernel-only Computer Vision challenge ever hosted on Kaggle, where the participants can only run their model within limited resources and are required to open-source their model (kernel) at the end of competition. We hope that this effort can further benefit the research community by increasing algorithm transparency and resource fairness.
\vspace{-10pt}

\section{Conclusions}
Artwork attribute recognition is a novel area of research in fine-grained visual classification (FGVC) domain. In this work, we propose a novel multi-attribute image dataset to enrich the assets of this research area. The iMet Collection 2019 dataset is the first high-quality artwork image dataset with research-grade attribute labels curated or verified by world-class museum experts.

As the attribute annotation work is an ongoing effort in the Met Open Access program, we plan to extend the iMet Collection challenge dataset with richer attributes covering more semantic groups, such as medium and size. In addition, we also plan to support more computer vision applications such as instance detection and segmentation, caption generation, and cross-modality retrieval in the future competitions.

{\small
\bibliographystyle{ieee_fullname}
\bibliography{egbib}
}

\end{document}